\documentclass[10pt,twocolumn,letterpaper]{article}

\usepackage{cvpr}
\usepackage{times}
\usepackage{epsfig}
\usepackage{graphicx}
\usepackage{amsmath}
\usepackage{amssymb}
\usepackage{subcaption}
\usepackage{makecell}
\usepackage{multirow}
\usepackage[export]{adjustbox}
\usepackage{mwe}
\usepackage{enumitem}
\usepackage{makecell}
\usepackage[numbers]{natbib}
\usepackage{pifont}
\usepackage{color, colortbl}

\usepackage[pagebackref=true,breaklinks=true,letterpaper=true,colorlinks,bookmarks=false]{hyperref}

\definecolor{Gray}{gray}{0.9}
\newcommand\blfootnote[1]{%
  \begingroup
  \renewcommand\thefootnote{}\footnote{#1}%
  \addtocounter{footnote}{-1}%
  \endgroup
}

\cvprfinalcopy

\def\cvprPaperID{6039} 

\ifcvprfinal\fi
\begin{document}

\title{A Transductive Approach for Video Object Segmentation}

\author{
	Yizhuo Zhang$^{2\star}$~~~~~~~~~~Zhirong Wu$^{1\star}$~~~~~~~~~~Houwen Peng$^{1}$~~~~~~~~~~Stephen Lin$^{1}$ \\
	$^1$Microsoft Research Asia~~~~~~~~~$^2$Carnegie Mellon University}

\maketitle

\begin{abstract}
Semi-supervised video object segmentation aims to separate a target object from a video sequence, given the mask in the first frame.
Most of current prevailing methods utilize information from additional modules trained in other domains like optical flow  and instance segmentation, and as a result they do not compete with other methods on common ground.
To address this issue, we propose a simple
yet strong transductive method, in which additional modules,
datasets, and dedicated architectural designs are not
needed.
Our method takes a label propagation approach where pixel labels are passed forward based on feature similarity in an embedding space.
Different from other propagation methods, ours diffuses temporal information in a holistic manner which take accounts of long-term object appearance. 
In addition, our method requires few additional computational overhead, and runs at a fast $\sim$37 fps speed. Our single model with a vanilla ResNet50 backbone achieves an overall score of $72.3\%$ on the DAVIS 2017 validation set and $63.1\%$ on the test set. This simple yet high performing and efficient method can serve as a solid baseline that facilitates future research.
Code and models are available at \url{https://github.com/microsoft/transductive-vos.pytorch}.
\end{abstract}

\newcommand{\HW}[1]{\textcolor[rgb]{0.0,0.0,1.0}{#1}}
\newcommand{\clear}[1]{\textcolor[rgb]{1.0,0.0,0.0}{#1}}

\section{Introduction}
\blfootnote{$^\star$Equal contribution. Work done when Yizhuo was an intern at MSRA.}

Video object segmentation addresses the problem of extracting 
object segments from a video sequence given the annotations in the starting frame.
This semi-supervised setting is challenging as 
it requires the system to generalize to various objects, deformations, and occlusions.
Nevertheless, video object segmentation has received considerable attention because of its broad practical applications in surveillance, self-driving cars, robotics, and video editing.

Despite the simplicity of the formulation, 
video object segmentation is closely related to many other visual problems,
such as instance segmentation~\cite{he2017mask}, object re-identification~\cite{farenzena2010person}, optical flow estimation~\cite{fischer2015flownet}, and object tracking~\cite{bernardin2008evaluating}. 
As these tasks share similar challenges with video object segmentation, previous efforts~\cite{li2018video,luiten2018premvos} attempt to transfer the modules trained for such tasks into video object segmentation pipeline. More specifically, optical flow and tracking encourage local dependencies by estimating displacements in nearby frames, while instance segmentation and object re-identification enforces global dependencies by learning invariances to large appearance changes.
The integration of such modules allows a significant performance improvement in video object segmentation.



\begin{figure}[t]
	\centering
	\includegraphics[width=0.95\linewidth]{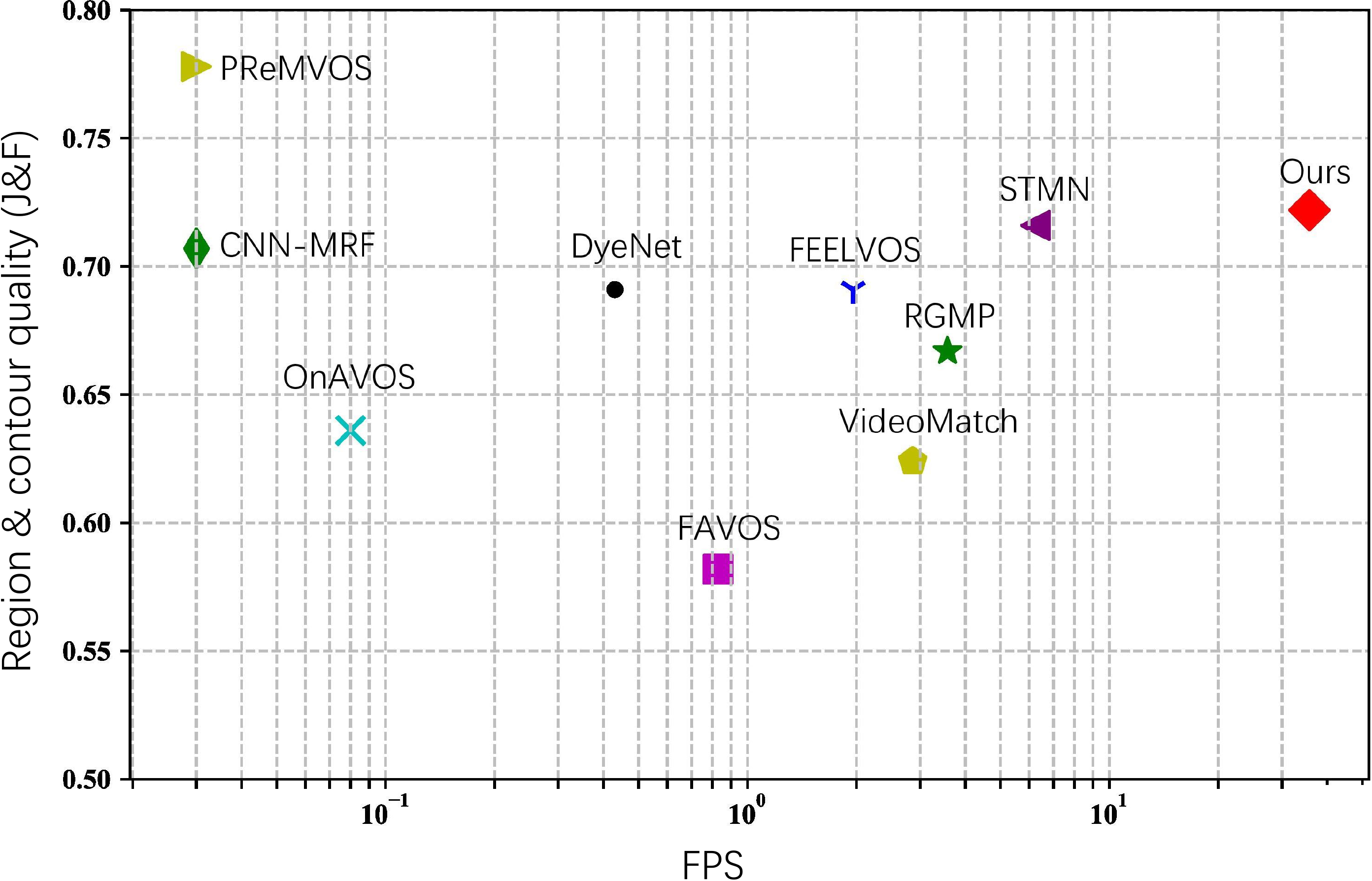}
	\caption{A comparison of performance and speed for semi-supervised video object segmentation methods on the DAVIS 2017 validation set. Ours performs comparably to the state-of-the-art methods, while running at an online speed ( $> 30$ fps).}
	\label{fig:speed_g}
\vspace{-1em}
\end{figure}

The idea of enforcing local and global
dependencies has been a central topic in general semi-supervised learning~\cite{zhou2004learning,fergus2009semi} (also known as transductive inference).
The basic assumptions are: 1) nearby samples
tend to have the same label and 2) samples
that lie on the same manifold should should have the same label.
The local and global dependencies describe a sufficiently smooth affinity distribution, so that label propagation on the unlabeled data gives reliable estimates.
Prior classical approaches that realize this idea include random walk~\cite{szummer2002partially}, graph-cut~\cite{blum2001learning} and spectral methods~\cite{belkin2002semi}.

This inspires us to explore a unified approach for semi-supervised video object segmentation without the integration of 
the modules derived from other domains.
We model the local dependency through a spatial prior and a motion prior.
It is based on the assumption that spatially nearby pixels 
are likely to have same labels and that temporally distant frames weakens the spatial continuity. On the other hand,
we model the global dependency through visual appearance, 
which is learned by convolutional neural networks on the training data.


The inference follows the regularization framework~\cite{zhou2004learning} which propagates labels in the constructed spatio-temporal dependency graph.
While label propagation algorithms have been explored in the recent literature for video object segmentation~\cite{voigtlaender2019feelvos,chen2018blazingly,hu2018videomatch,shin2017pixel,perazzi2017learning}, the manner in which they learn and propagate affinity is sparse and local, i.e., learning pixel affinities either between adjacent frames or between the first frame and a distant frame.
We observe that there exists much smooth unlabeled structure in a temporal volume that these methods do not exploit. 
This may cause failures when handling deformations and occlusions. In contrast, our 
label propagation approach attempts to capture all frames which span the video sequence from the first frame to the frame preceding the current frame.
To limit the computational overhead, sampling is performed densely within the recent history and sparsely in the more distant history, yielding a model that accounts for object appearance variation while reducing temporal redundancy.


In its implementation, our model does not rely on any other task modules, additional datasets, nor dedicated architectural designs beyond a pretrained ResNet-50 model from the ImageNet model zoo~\cite{he2016deep}.
During inference, per-frame prediction involves only a feed-forward pass through the base network plus an inner product with the prediction history.
Thus the inference is fast and also not affected by the number of objects.
Experimentally, our model runs at a frame rate of $37$ per second, achieving an overall score of $72.3\%$ on Davis 2017 validation set, as well as $63.1\%$ on Davis 2017 test set.
Our model also achieves a competitive overall score of $67.8\%$ on the recent Youtube-VOS validation set.
Our method is competitive to current prevailing methods while being substantially simpler and faster. We hope the model can serve as a simple baseline for future works.

\section{Related Work}

\begin{figure*}[t]
	\centering
	\includegraphics[width=0.95\linewidth]{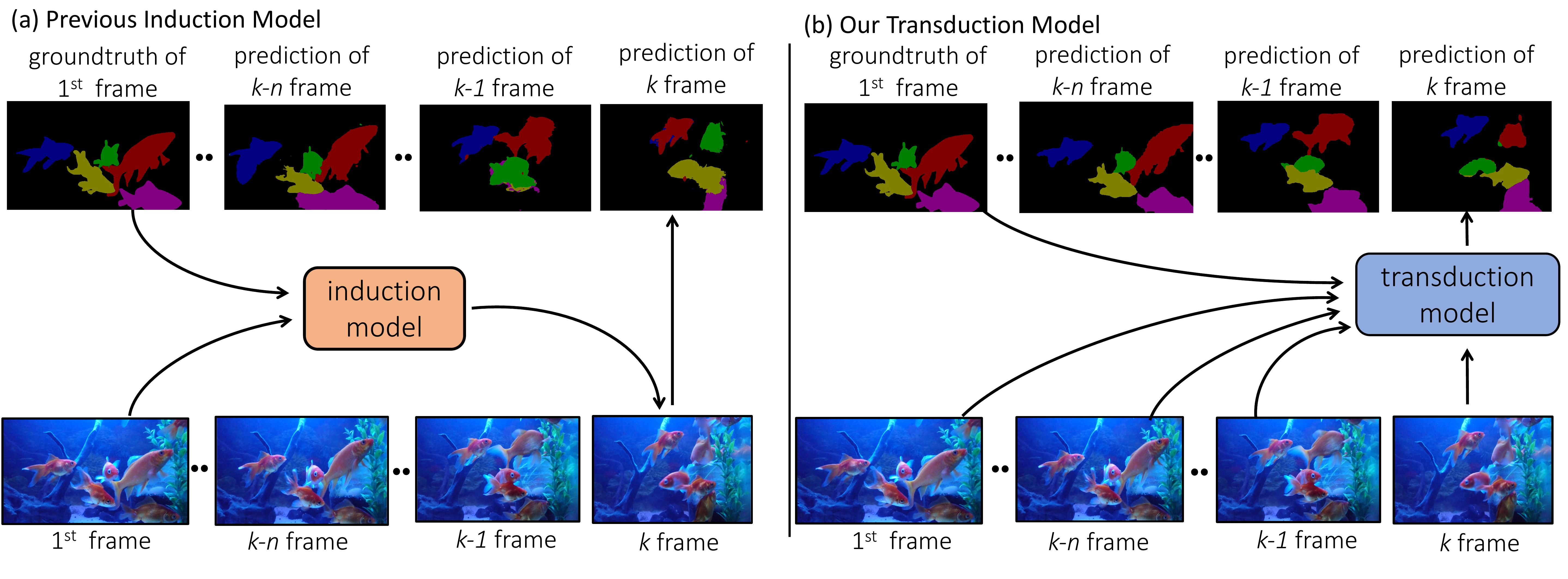}
	\caption{We pose video object segmentation from a transductive inference perspective, where dense long-term similarity dependencies are constructed to discover structures in the spatio-temporal volume. a) Previous induction model transfers knowledge from the first frame to other frames. b) Our transduction model considers holistic dependencies in the unlabeled spatio-temporal volume for joint inference.}
	\label{fig:pipeline}
\end{figure*}

We review related work on video object segmentation in the semi-supervised setting. For an overview of unsupervised and interactive video object segmentation, we refer readers to other papers~\cite{faktor2014video,fan2015jumpcut,bai2009video,tsai2016video,li2018instance, li2018unsupervised}.

\vspace{3pt}
\noindent \textbf{Single frame models. }
In the past few years, methods with leading performance have been based on finetuning the model on the single annotated frame and performing inference on individual test frames.
These methods basically learn an objectness prior and spatial continuity without considering temporal information.
The convolutional neural network architecture plays an important role for finetuning on a single frame to be effective.
OSVOS~\cite{caelles2017one} is the pioneering work in this direction.
Lucid~\cite{khoreva2018lucid} seeks to augment the data specifically for each video from only a single frame of ground truth.
OnAVOS~\cite{voigtlaender2017online} mines confident regions in the testing sequence to augment the training data.
The later work of OSVOS-S~\cite{maninis2017video} integrates semantic information from an instance segmentation model to boost performance.
PReMVOS~\cite{luiten2018premvos}, CNN-MRF~\cite{bao2018cnn}, and DyeNet~\cite{li2018video} also build on top of single frame models.

The effectiveness of single frame models demonstrates that optimizing a domain-specific spatial smoothness term greatly enhances performance. However, finetuning via gradient descent generally takes tens of seconds per video, which can make it impractical for many applications.

\vspace{3pt}

\noindent \textbf{Propagation-based models. } 
Propagation-based methods embed image pixels into a feature space and utilize pixel similarity in the feature space to guide label propagation.
In methods such as VideoMatch~\cite{hu2018videomatch,chen2018blazingly}, only pixels in the first frame are used for reference in computing pixel similarity. 
Since no finetuning at run time is involved, propagation-based models run much faster than the aforementioned single frame models, but the lack of domain-specific finetuning leads to performance that is much worse.
Later works~\cite{shin2017pixel,wug2018fast,perazzi2017learning,voigtlaender2019feelvos} explore adding the preceding frame to the first frame as reference, which significantly improves performance and leads to greater temporal smoothness. However, this local and sparse propagation scheme suffers from the drifting problem~\cite{grabner2008semi}.

\vspace{3pt}

\noindent \textbf{Long-range spatio-temporal models. }
There are two lines of work which attempt to optimize over a dense long-range spatio-temporal volume.
The first~\cite{wug2018fast, hu2017maskrnn} builds a recurrent neural network which uses the estimate from the previous frame to predict the object segmentation in the current frame. The whole model is learned via backpropagation through time. However, such models are sensitive to estimation errors in the previous frame.

The second direction is based on graphical models~\cite{grundmann2010efficient, tsai2012motion,lee2011key,zhang2013video,chandra2018deep,marki2016bilateral} (i.e., Markov Random Fields) defined over the spatio-temporal domain.
These works were popular prior to deep learning, and employed edge potentials defined by handcrafted features such as SIFT.
The models are computationally expensive and no longer competitive to learning-based methods.


\vspace{3pt}

\noindent \textbf{Relation to other vision problems. }
As the above methods suggest, video object segmentation is closely related to a variety of computer vision problems such as instance segmentation, object re-identification, and optical flow estimation and tracking. Many recent methods integrate components for these other tasks into the video object segmentation pipeline.
For example, OSVOS-S~\cite{maninis2017video} includes a instance segmentation module; PReMVOS~\cite{luiten2018premvos} and DyeNet~\cite{li2018video} incorporate a object re-identification module; CNN-MRF~\cite{bao2018cnn}, MaskTrack~\cite{perazzi2017learning} and MaskRNN~\cite{hu2017maskrnn} rely on optical flow estimation.
The integration of other modules heavily depends on transfer learning from other datasets. 
Though performance improvement is observed, it usually involves
further complications.
For example, 
instance segmentation becomes less useful when the video encounters a new object category which is not present in instance segmentation model.
Optical flow~\cite{fischer2015flownet} suffers from occlusions which would mislead label propagation. 
\vspace{3pt}

\vspace{3pt}

\noindent \textbf{Most relevant works. } Space-time memory network (STM)~\cite{oh2019video} is a significant work and most 
similar with ours.
Ours is developed independently from 
STM, 
while STM is published earlier than ours. The insight that exploiting dense long-term information is similar. 
However, the transductive framework in the proposed approach, which stems from the classical semi-supervised learning, brings theoretical foundations to video object segmentation. Moreover, in the implementation, ours is much simpler and more efficient which does not require additional datasets, and infers all objects simultaneously. 

\section{Approach}
In contrast to much prior work on finetuning a model on a single annotated frame or transferring knowledge from other related tasks,
our approach focuses on fully exploiting the unlabeled structure in a video sequence.
This enables us to build a simple model that is both strong in performance and fast in inference.

We first describe a generic semi-supervised classification framework~\cite{zhou2004learning} and then adapt it to online video object segmentation in a manner that follows our ideas. 

\subsection{A Transductive Inference Framework}
Let us first consider a general semi-supervised classification problem.
Suppose that we have a dataset $\mathcal{D} = \{(x_1, y_1), (x_2, y_2), (x_l, y_l), x_{l+1}, ...,  x_n \}$, which contains $l$ labeled data pairs and $n-l$ unlabeled data points.
The task is to infer the labels $\{\hat{y}_i\}_{i=l+1}^n$ for the unlabeled data $\{x_{l+1}, ... ,x_n\}$ based on all the observation $\mathcal{D}$. 
Inference of the unlabeled data is formulated in prior work~\cite{zhou2004learning} as a transductive regularization framework,
\begin{equation}
\label{generic}
\mathcal{Q}(\hat{\bf{y}}) =   \sum_{i,j}^n w_{ij} || \frac{\hat{y}_i}{\sqrt{d_i}}- \frac{\hat{y}_j}{\sqrt{d_j}}||^2  + \mu \sum_{i=1}^l ||\hat{y}_i - y_i||^2 ,
\end{equation}
where $w_{ij}$ encodes the similarity between data points $(x_i, x_j)$,
and $d_i$ denotes the degree $d_i = \sum_j w_{ij}$ for pixel $i$. 
The first term is a smoothness constraint that 
enforces similar points to have identical 
labels. The second term is a fitting constraint, which penalizes solutions that deviate from the initial observations. The parameter $\mu$ balances these two terms. Semi-supervised classification amounts to solve the following optimization problem,
\begin{equation}
\label{obj}
    \hat{\bf{y}} = \text{argmin }\mathcal{Q}(\bf{y}).
\end{equation}

It is shown~\cite{zhou2004learning} that the above energy minimization problem can be solved by iterative algorithm as follows. 
Let ${\bf{S}} = {\bf{D}}^{-1/2}{\bf{WD}}^{-1/2}$  
be the normalized similarity matrix constructed from $w_{ij}$.
Iteratively solve for $\hat{\bf{y}}(k)$ until convergence, as~\footnote{Note that there exists a closed-form solution to the Eqn~\ref{obj} shown in~\cite{zhou2004learning}. However, this requires the inverse of matrix $S$ which is often computationally demanding when $S$ is a large matrix.} 
\begin{equation}
\label{prop}
\hat{\bf{y}}(k+1) = \alpha {\bf{S}} \hat{\bf{y}}(k) + (1-\alpha) {\bf{y}}(0),
\end{equation}
where $\alpha = \mu / (\mu + 1)$, and ${\bf{y}}(0)=[y_1, y_2,...,y_n]^T$ is the initial observation of the label clamped with supervised labels.
The typical value of $\alpha$ is $0.99$. The power of this transduction model comes from the globalized model it builds over the dense structures in the unlabeled data.

\subsection{Online Video Object Segmentation}

Based on this general framework~\cite{zhou2004learning}, we build a transductive model for semi-supervised video object segmentation that accounts for dense long-range interactions.

This gives rise to three challenges.
First, video frames stream sequentially, so the model must work in an online fashion, where the inference of one frame should not depend on future frames. 
Second, the number of pixels in one video can scale into the tens of millions. A similarity matrix over all the pixels would thus be intractable to compute.
Third, an effective similarity measure $W$ needs to be learned between pixels in a video sequence.

For the algorithm to run online,
it is assumed that the predictions on all prior frames have been determined
when the current frame $t$ arrives. 
We therefore approximate the Eqn~\ref{prop} by expanding the inference procedure through time,
\begin{equation}
\label{propt}
\hat{\bf{y}}(t+1) = {\bf{S}}_{1:t \to t+1} \hat{\bf{y}}(t).
\end{equation}
${\bf{S}}_{1:t \to t+1}$ represents the similarity matrix $\bf S$ that is only constructed between pixels up to the $t$-th frame and the pixels in the $t+1$-th frame.
Since no labels are provided beyond the first frame, the prior term $\bf y(0)$ is omitted for the frame $t+1$. 

For time $t+1$,
the above propagation procedure is equivantly minimizing a set of smoothness terms in the spatio-temporal volume,
\begin{equation}
\label{timet}
\mathcal{Q}^{t+1}(\hat{{\bf y}}) =  \sum_{i}  \sum_{j} w_{ij}|| \frac{\hat{y}_i}{\sqrt{d_i}}- \frac{\hat{y}_j}{\sqrt{d_j}}||^2 ,
\end{equation}
where $i$ indexes the pixels at the target time $t+1$, $j$ indexes the pixels in all frames prior to and including time $t$. 

\subsection{Label Propagation}
Given the annotations on the starting frame of a video, we process the remaining frames sequentially, propagating labels to each frame based on Eqn.~\ref{propt}. 
The quality of video object segmentation heavily depends on the similarity metric $\bf{S}$, whose core component is the the affinity matrix $\bf W$.
 
\vspace{2pt}
\textbf{Similarity metric. } In order to build a smooth classification function, the similarity metric should account for global high-level semantics and local low-level spatial continuity. Our similarity measure $w_{ij}$ includes an appearance term and a spatial term,  
\begin{equation}
w_{ij} = \text{exp}({f}^T_i{f}_j) \cdot \text{exp} (-\frac{||\text{loc}(i) - \text{loc}(j)||^2}{\sigma^2}),
\end{equation}
where ${f}_i, {f}_j$ are the feature embeddings for pixels $p_i, p_j$ through a convolutional neural network. 
$\text{loc}(i)$ is the spatial location of pixel $i$.
The spatial term is controlled by a locality parameter $\sigma$.
Learning of the appearance model is described in the next section.

\vspace{2pt}
\textbf{Frame sampling. } 
Computing a similarity matrix $\bf S$ over all the previous frames is computationally infeasible, as long videos can span hundreds of frames or more.
Inspired by Temporal Segment Networks~\cite{wang2016temporal}, we sample a small number of frames in the observance of the temporal redundancy in videos. 

\begin{figure}[t]
	\centering
	\includegraphics[width=1.\linewidth]{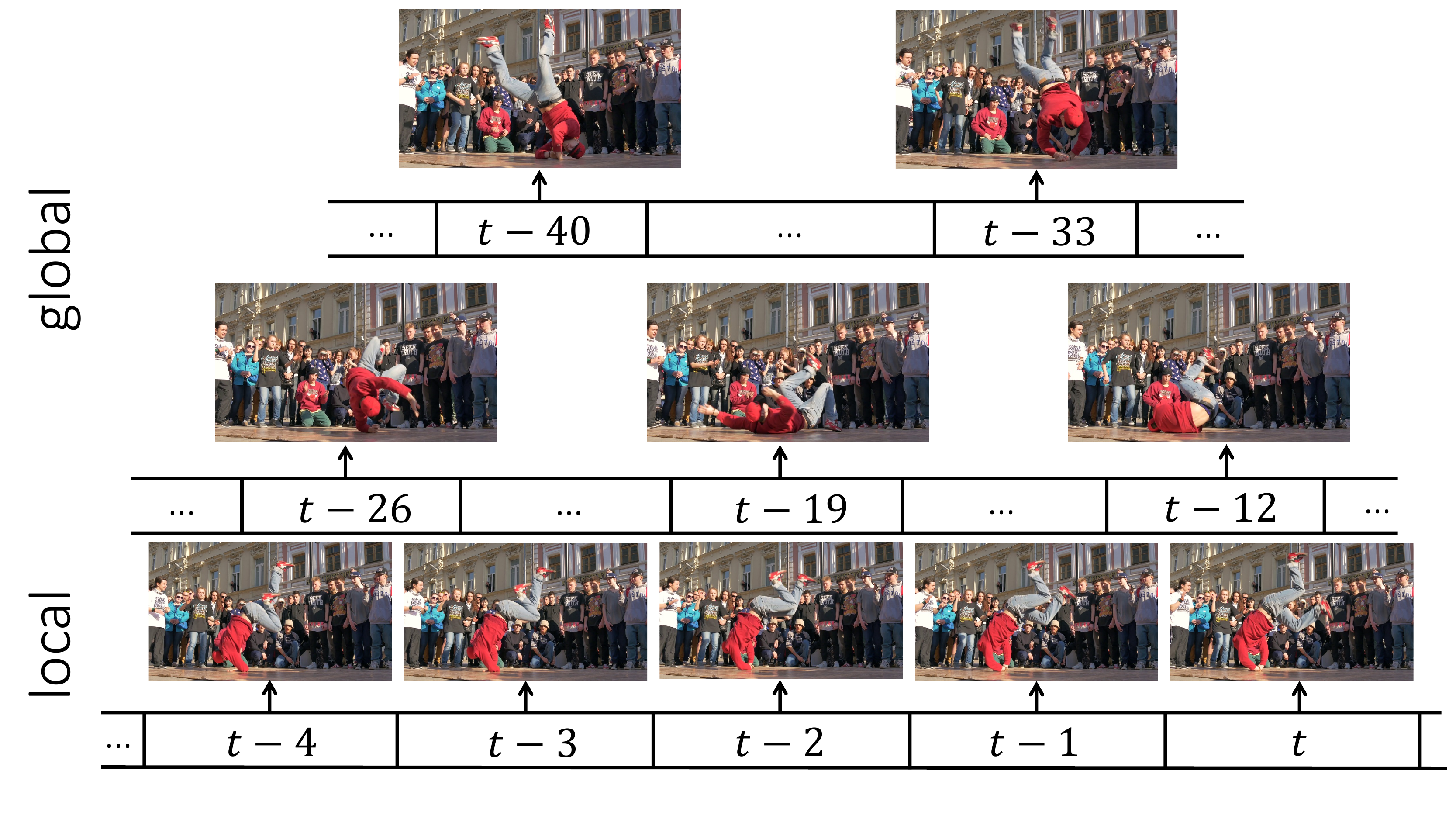}
	\caption{Sampling strategy for label propagation. We sample densely in the recent history, and more sparsely in the distant history.  }
	\label{fig:sampling}
\end{figure}

Specifically, as in Figure~\ref{fig:sampling}, we sample a total of 9 frames from the preceding 40 frames: the 4 consecutive frames before the target frame to model the local motion, and 5 more frames sparsely sampled from the remaining 36 frames to model long-term interactions.
We find this sampling strategy to strike a good balance between efficiency and effectiveness.
Detailed ablations about the choice of frame sampling are presented in the experiments. 

\vspace{2pt}
\textbf{A simple motion prior. } 
Pixels that are more distant in the temporal domain have weaker spatial dependencies.
To integrate this knowledge, we use a simple motion prior where
a smaller $\sigma = 8$ is used when the temporal references are sampled locally and densely, and a larger $\sigma = 21$ is employed when the reference frames are distant.
We find this simple motion model to be effective for finding long-term dependencies.

\begin{table}[t]
	\setlength{\tabcolsep}{1.3pt}
	\centering
	\small
	\begin{tabular}{cccccc}
		\Xhline{2\arrayrulewidth} 
		Methods & Architecture & Optical & Proposal & Tracking & Re-ID  \\
		\Xhline{2\arrayrulewidth} 
		DyeNet~\cite{li2018video} & ResNet 101 & \ding{51} & \ding{51} &  \ding{51} &  \ding{51} \\
		CNN-MRF~\cite{bao2018cnn} & Deeplab & \ding{51} & \ding{55} & \ding{55} & \ding{55} \\
		PReMVOS~\cite{luiten2018premvos} & Deeplab-V3+ & \ding{51} & \ding{51} & \ding{51} & \ding{51} \\
		\hline
		
		FEELVOS~\cite{voigtlaender2019feelvos} & Deeplab-V3+ & \ding{55} & \ding{55} & \ding{55} & \ding{55} \\
		STM~\cite{oh2019video}  &  2$\times$ResNet-50 & \ding{55} & \ding{55} & \ding{55} & \ding{55}\\
		\rowcolor{Gray}
		TVOS (ours) &  ResNet-50 & \ding{55} & \ding{55} & \ding{55} & \ding{55}\\
		\Xhline{2\arrayrulewidth}
	\end{tabular}
	\caption{A brief overview of leading VOS methods with dependent modules for other related vision tasks.}
	\label{table:methods}
\end{table}

\subsection{Learning the appearance embedding}
We learn the appearance embedding in a data-driven fashion using a 2D convolutional neural network.
The embedding aims to capture both short-term and long-term variations due to motion, scale and deformations.
The embedding is learned from the training data in which each frame from the video is annotated with the segmented object and the object identity. 

Given a target pixel $x_i$ and we consider all pixels in the prior frames as references. 
Denote $f_i$ and $f_j$ the feature embeddings for pixel $x_i$ and a reference pixel $x_j$.
Then the predicted label $\hat{y_i}$ of $x_i$ is given by
\begin{equation}
\hat{y_i} = \sum_j \frac{ \text{exp}( {f}_i^T{f}_j) }{\sum_k \text{exp}( {f}_i^T{f}_k) }\cdot y_j ,
\end{equation}
where the reference indexes $j,k$
span the temporal history before the current frame. 
We show detailed ablations on how sampling the historical frames affects the learning quality.

We optimize the embedding via a standard cross-entropy loss on all pixels in the target frame,
\begin{equation}
\mathcal{L} = -\sum_i \text{log} P(\hat{y}_i = y_i | x_i).
\end{equation}

\subsection{Implementation Details}

We use a ResNet-50 to train the embedding model.
The convolution stride of the third and the fourth residual blocks is set to 1 to maintain a high-resolution output.
We add one additional $1\times 1$ convolutional layer to project the feature to a final embedding of 256 dimensions. 
The embedding model produces a feature with a total stride of 8.

During training, we take the pretrained weights from the ImageNet model zoo, and finetune the model on the Davis 2017~\cite{davis2017} training set for 240 epochs and Youtube-VOS~\cite{xu2018youtube} for 30 epochs.
We apply the standard augmentations of random flipping and random cropping of size $256\times 256$ on the input images.
We use a SGD solver with an initial learning rate of $0.02$ and a cosine annealing scheduler. 
The optimization takes 16 hours on 4 Tesla P100 GPUs, with a batch size of 16, each containing 10 snippets from a video sequence.

During tracking, we extract features at the original image resolution of 480p. The results of each video frame are predicted sequentially online.

\begin{figure}[t]
	\centering
	\includegraphics[width=0.8\linewidth]{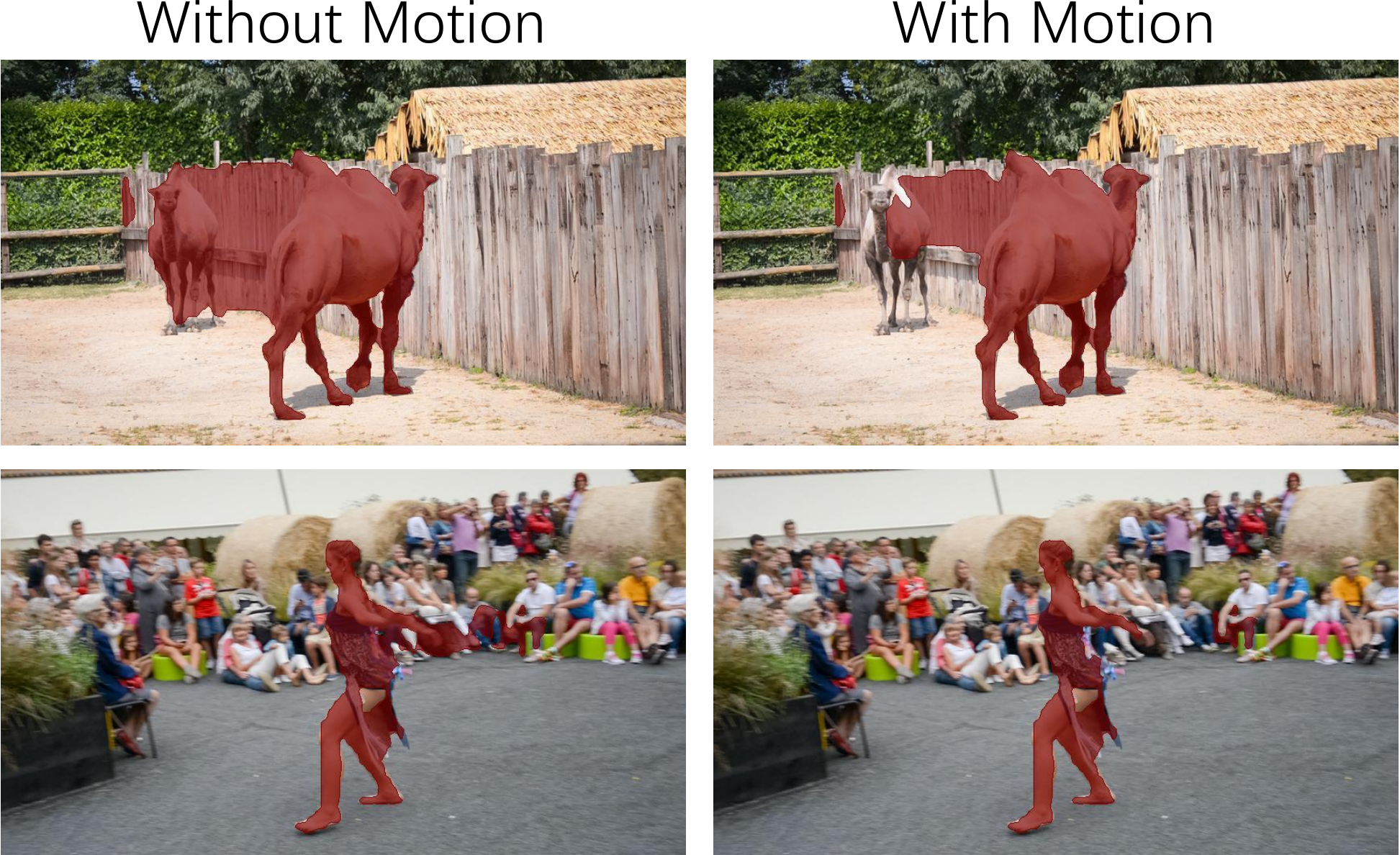}
	\caption{The effect of our simple motion model. Distant frames have weaker spatial priors on the location of objects, thus reducing the drifting problem. }
	\label{fig:motion}
\end{figure}

\section{Results}

In this section, we first describe our experimental settings and datasets. 
Then we show detailed ablations on how the transductive approach takes advantage of unlabeled structures in the temporal sequence to significantly improve the performance.
Results are conducted on various datasets to compare with the state-of-the-art.
Finally, we discuss temporal stability and the relationship to optical flow. Our method is abbreviated as TVOS in the result tables.

\begin{figure*}[t]
	\centering
	\includegraphics[width=0.98\linewidth]{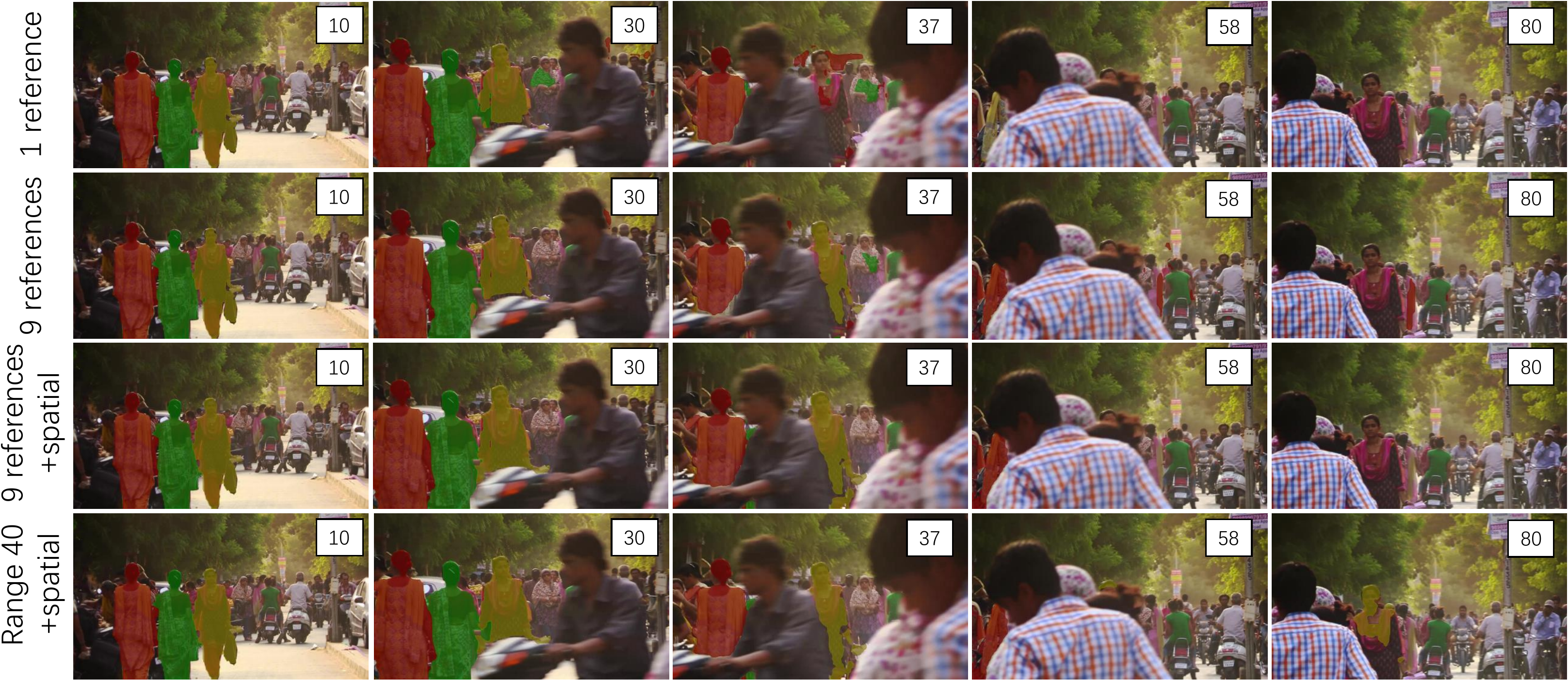}
	\caption{Using dense long-range dependencies improves the tracking performance. The spatial term smooths the object boundaries, while long-term dependencies up to 40 frames help to re-detect objects.}
	\label{fig:temporal}
\end{figure*}

\subsection{Experimental Setup}

\textbf{Datasets.}
We evaluate our method on the Davis 2017~\cite{davis2017}, and Youtube-VOS~\cite{xu2018youtube} datasets.
Our model is trained on the respective training set and evaluated on the validation set.
For Davis 2017, we also train our model on the combined train-val set, and submit the results on the testing set on the evaluation server.

Davis 2017 contains 150 video sequences and it involves multiple objects with drastic deformations, heavy and prolonged occlusions, and very fast motions.
High-definition annotations are available for all frames in the training sequences.
Youtube-VOS is the largest dataset for this task to-date, containing 4453 training sequences and 474 validation sequences.
It captures a comprehensive collection of $94$ daily object categories.
However, the frame rate of the video is much lower than videos in Davis (5 fps compared to 24 fps).

\begin{table}[t]
	\setlength{\tabcolsep}{1.8pt}
	\centering
	\small
	\begin{tabular}{c|cccccc}
		\Xhline{2\arrayrulewidth} 
		train / tracking & \begin{tabular}{@{}c@{}}1 \\ frame\end{tabular} & \begin{tabular}{@{}c@{}}3 \\ frames \end{tabular} & \begin{tabular}{@{}c@{}}9 \\ frames\end{tabular} & \begin{tabular}{@{}c@{}}uniform \\ sample \end{tabular} & \begin{tabular}{@{}c@{}}sparse \\ sample\end{tabular}& 
		\begin{tabular}{@{}c@{}}sparse + \\ motion \end{tabular} \\ 
		\hline
		1 frame  & 55.8  & 60.4 & 63.4 & 63.8 & 64.0 & 64.3 \\
		3 frames & 56.0 & 61.4 & 65.4 & 65.5 & 66.1 & 66.7\\
		9 frames & 60.7 & 63.4 & 68.6 & 68.6 & 69.0 & 69.9 \\
		uniform sample & 55.8 & 60.2 & 64.4 & 65.0 & 65.1 & 65.3 \\
		sparse sample & 59.9 & 62.9 & 66.2 & 67.2 & 68.5 & 68.6\\
		\Xhline{2\arrayrulewidth}
		Supervised & 47.5 & 52.2 & 53.8 & 54.0 & 54.5 & 54.8\\
		InstDisc\cite{wu2018unsupervised} & 42.4 & 47.3 & 51.3 & 51.3 & 52.1 & 52.2\\
		MoCo\cite{he2019momentum} & 43.5 & 48.7 & 53.0 & 53.2 & 53.8 & 54.0\\
		\Xhline{2\arrayrulewidth}
	\end{tabular}
	\caption{Ablation study on the range of temporal dependencies and the simple motion component. The mean $J$ measure on the Davis 2017 validation set is reported. See text for details.}
	\label{table:ablation}
\end{table}

\vspace{2pt}
\textbf{Evaluation metrics.}
We use the standard evaluation metric of mean intersection over union (mIoU), averaged across objects and summed over all frames.
The mIoU is evaluated on both the full objects (J measure) and only on the object boundaries (F measure).
The global metric (G measure) is the average of the J and F measures.
Youtube-VOS also includes a separate measure of seen objects and unseen objects to measure the generalization ability.
In Section~\ref{stable}, we provide a discussion of temporal stability.

\subsection{Ablation Study}

\textbf{Dense local and global dependencies. } 
While most prior works focus on optimizing single frame models, the key idea of this paper is to build dense long-term models over 
the spatio-temporal volume. 
In Table~\ref{table:ablation}, we summarize the effect of such long-term potentials which capture both local and global dependency.
Each row is an appearance embedding model trained with different reference frame sampling strategies.
Each column corresponds to a tracking sampling strategy.
We study the following settings: one reference frame preceding the target frame, 3 consecutive frames preceding the target frame, 9 consecutive frames preceding the target frame, uniform sampling of 9 frames in the preceding 40 frames, and our sparse sampling of 9 frames in the preceding 40 frames as in Figure~\ref{fig:sampling}.
We find that tracking over a longer term generally improves the performance,
and denser sampling near the target frame is helpful.
For learning the appearance embedding, training with 9 consecutive frames produces the best results, while longer ranges do not always lead to improvements.
This may be due to very long ranges covering almost the entire video reduces the variations in the dataset, which leads to worse generalization for training.

In Figure~\ref{fig:temporal}, we show some qualitative examples for long range tracking. Using 9 consecutive frames yields more stable predictions than using only the previous frame. Adding the spatial term smooths the object boundaries. A long range of 40 frames enables the model to re-detect objects after heavy occlusions.  

\textbf{Transferred representations. } In the last rows of Table~\ref{table:ablation}, we also test the tracking performance for models pretrained on ImageNet but without further training on the DAVIS dataset. 
The transferred ImageNet model obtains a mean $J$ measure of $54.8\%$, which is actually better than some prior methods~\cite{yang2018efficient,cheng2018fast} trained with additional Davis data.
Also, even an unsupervised pretrained model on images obtains performance competitive to network modulation~\cite{yang2018efficient} using our transductive inference algorithm.
Two recent unsupervised pretrained models on ImageNet are investigated~\cite{wu2018unsupervised,he2019momentum}.
Since no domain-specific training is involved for the appearance embedding, 
the evaluation of transferred representations clearly validates the effectiveness of dense long-term modeling.

\vspace{2pt}
\textbf{The simple motion prior. } 
As a weak spatial prior for modeling the dependency between distant frames,
our simple motion model reduces noise from the model predictions and leads to about $1\%$ improvement.
Figure~\ref{fig:motion} displays two concrete examples.
More complicated motion models, such as a linear motion model~\cite{andriyenko2011multi}, may be even more effective.

\begin{table}[t]
	\setlength{\tabcolsep}{6pt}
	\centering
	\small
	\begin{tabular}{cccccc}
		\Xhline{2\arrayrulewidth} 
		Methods & FT & $\mathcal{J}$ & $\mathcal{F}$ & $\mathcal{J}\&\mathcal{F}$ & Speed  \\
		\Xhline{2\arrayrulewidth} 
		
		OnAVOS~\cite{voigtlaender2017online} & \ding{51} & 61.0 & 66.1 & 63.6 & 0.08 \\
		DyeNet~\cite{li2018video} & \ding{51} & 67.3 & 71.0 & 69.1 & 0.43 \\
		CNN-MRF~\cite{bao2018cnn} & \ding{51} & 67.2 & 74.2 & 70.7 &  0.03 \\
		PReMVOS~\cite{luiten2018premvos} & \ding{51} & \textbf{73.9} & \textbf{81.7} & \textbf{77.8} & \textbf{0.03} \\
			\Xhline{2\arrayrulewidth}
			Modulation~\cite{yang2018efficient} & \ding{55}  & 52.5 & 57.1 & 54.8 & 3.57 \\
		FAVOS~\cite{cheng2018fast} &\ding{55}  & 54.6 & 61.8 & 58.2 & 0.83 \\
		VideoMatch~\cite{hu2018videomatch} &\ding{55} & 56.5 & 68.2 & 62.4 & 2.86 \\
		RGMP~\cite{wug2018fast} & \ding{55}& 64.8 & 68.8 & 66.7 & 3.57  \\
		FEELVOS~\cite{voigtlaender2019feelvos} &\ding{55} &  65.9 & 72.3 & 69.1 & 1.96 \\
		STM~\cite{oh2019video} &\ding{55} &  69.2 & 74.0 & 71.6 & 6.25 \\
		STM~\cite{oh2019video}+Pretrain &\ding{55} & 81.7 & 79.2 & 84.3 & 6.25\\
		\rowcolor{Gray}
		TVOS & \ding{55} & \textbf{69.9} & \textbf{74.7} & \textbf{72.3} & \textbf{37} \\

		\Xhline{2\arrayrulewidth}
	\end{tabular}
	\caption{Quantitative evaluation on the Davis 2017 validation set. FT denotes methods that perform online training.}
	\label{table:davis17-val}
\end{table}

\subsection{Quantitative Results}

In Table~\ref{table:methods}, we first give a brief overview of the current leading methods, including those that use first-frame finetuning (CNN-MRF~\cite{bao2018cnn}, DyeNet~\cite{li2018video}, PReMVOS~\cite{luiten2018premvos}) and those that do not (FEELVOS~\cite{voigtlaender2019feelvos}, STM~\cite{oh2019video} and our TVOS).
For DyeNet and PReMVOS, their sub-modules are learned on dedicated datasets such as optical flow on Flying Chairs, object proposal on MSCOCO, and object segmentation on PASCAL VOC.
Since Davis is much smaller than the large-scale datasets, it remains unknown how much of the gains can be attributed to knowledge transfer or to the methods themselves.
Therefore, the mentioned methods are not directly comparable with our method.
FEELVOS, STM and ours are much simpler, as they do not rely on additional modules for this problem. STM additionally requires heavy pretraining on large-scale image datasets.

It is also important to note that for PreMVOS, DyeNet, CNN-MRF, they are not able to run tracking \textbf{in an online fashion}.
They use information from future frames to stabilize prediction for the target frame. Also, instead of using the first frame from the given video for training, they use the first frames from the entire test-dev set for training. 
Propagation-based methods are able to track objects sequentially online.

\vspace{2pt}
\textbf{DAVIS 2017.}
We summarize our results on the Davis 2017 validation set in Table~\ref{table:davis17-val}, and on the Davis 2017 test-dev set in Table~\ref{table:davis17-test}.
On the validation set,
our method performs slightly better than STM~\cite{oh2019video} under the same amount of training data, while surpassing other propagation-based methods which do not need fine-tuning, by $4\%$ for mean $J$ and $3\%$ for mean $J\& F$. 
In comparison to finetuning based methods, 
our TVOS also outperforms DyeNet and CNN-MRF by $2\%$ while being significantly simpler and faster.

We train our model on the combined training and validation set for evaluating on the test-dev set. We find that there is a large gap of distribution between the Davis 2017 test-dev and validation sets.
Heavy and prolonged occlusions among objects belonging to the same category are more frequent in the test-dev, which favors methods with re-identification modules.
As a result, we are $4-5\%$ lower than DyeNet and CNN-MRF on the test-dev set.
FEELVOS is even more negatively affected, performing $8\%$ lower than ours in terms of mean $J\&F$.
STM~\cite{oh2019video} does not provide an evaluation on the test set.

\begin{table}[t]
	\setlength{\tabcolsep}{7pt}
	\centering
	\small
	\begin{tabular}{cccccc}
		\Xhline{2\arrayrulewidth} 
		Methods & FT & $\mathcal{J}$ & $\mathcal{F}$ & $\mathcal{J}\&\mathcal{F}$ & Speed  \\
			\Xhline{2\arrayrulewidth} 
				OnAVOS~\cite{voigtlaender2017online} & \ding{51} & 53.4 & 59.6 & 56.5 & 0.08 \\
		DyeNet~\cite{li2018video} & \ding{51} & 65.8 & 70.5 & 68.2 & 0.43 \\
		CNN-MRF~\cite{bao2018cnn} & \ding{51} & 64.5 & 70.5 & 67.5 & 0.02 \\
		PReMVOS~\cite{luiten2018premvos} & \ding{51} & \textbf{67.5} & \textbf{75.7} & \textbf{71.6} & \textbf{0.02} \\
		\Xhline{2\arrayrulewidth}

		RGMP~\cite{wug2018fast} &\ding{55} & 51.4 & 54.4 & 52.9 & 2.38  \\
		FEELVOS~\cite{voigtlaender2019feelvos} &\ding{55} & 51.2 & 57.5 & 54.4 & 1.96 \\
		\rowcolor{Gray}
		TVOS & \ding{55}& \textbf{58.8} & \textbf{67.4} & \textbf{63.1} & \textbf{37} \\
		\Xhline{2\arrayrulewidth}

	\end{tabular}
	\caption{Quantitative evaluation on the Davis 2017 test-dev set. FT denotes methods that perform online training.}
	\label{table:davis17-test}
\end{table}

\begin{table}[t]
	\setlength{\tabcolsep}{6pt}
	\centering
	\small
	\begin{tabular}{cccccc}
		\Xhline{2\arrayrulewidth} 
		\multirow{2}{*}{Methods} & \multirow{2}{*}{Overall} & \multicolumn{2}{c}{Seen} & \multicolumn{2}{c}{Unseen} \\
		\cline{3-6}
		& &  $\mathcal{J}$  & $\mathcal{F}$  & $\mathcal{J}$  & $\mathcal{F}$ \\
		\Xhline{2\arrayrulewidth}
		RGMP~\cite{wug2018fast}  & 53.8 & 59.5 & - & 45.2 & - \\
		OnAVOS~\cite{voigtlaender2017online} & 55.2 & 60.1 & 62.7 & 46.6 & 51.4\\
		RVOS~\cite{ventura2019rvos} & 56.8 & 63.6 & 67.2 &  45.5 & 51.0\\
		OSVOS~\cite{caelles2017one} & 58.8 & 59.8 & 60.5 & 54.2 & 60.7\\
		S2S~\cite{xu2018youtube} & 64.4 & 71.0 & 70.0 & 55.5 & 61.2\\
		PreMVOS~\cite{luiten2018premvos} & 66.9 & 71.4 & 75.9 & 56.5 & 63.7\\
		STM~\cite{oh2019video}+Pretrain & 79.4 & 79.7 & 84.2 & 72.8 & 80.9 \\
		\rowcolor{Gray}
		TVOS & 67.8 & 67.1 & 69.4 & 63.0 & 71.6\\
		\rowcolor{Gray}
		TVOS (from DAVIS)  & 67.4 & 66.7 & 69.8 & 62.5 & 70.6\\

		\Xhline{2\arrayrulewidth}
	\end{tabular}
	\caption{Quantitative evaluation on the Youtube-VOS validation set.}
	\label{table:youtube}
\end{table}

\begin{figure*}[t]
	\centering
	\includegraphics[width=0.98\linewidth]{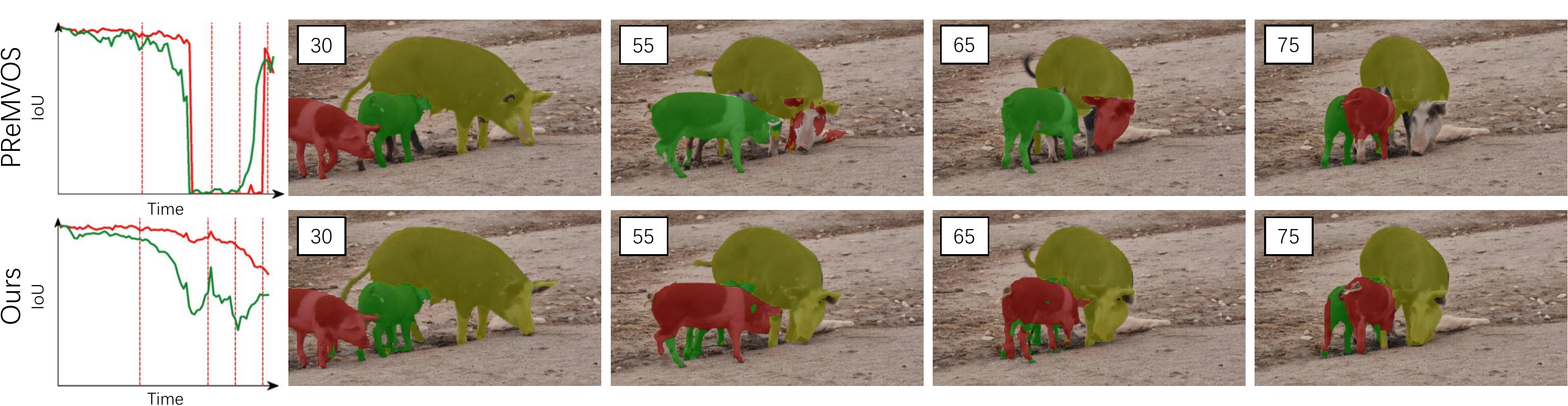}
	\caption{Per-frame IoU over time for PreMVOS and our method on a example video sequences from the DAVIS validation set. PreMVOS switches object identities frequently, while our predictions are temporally smooth. The color of each IoU curve matches its corresponding object segment. }
	\label{fig:stable}
\end{figure*}

\begin{figure}[t]
	\centering
	\includegraphics[width=0.9\linewidth]{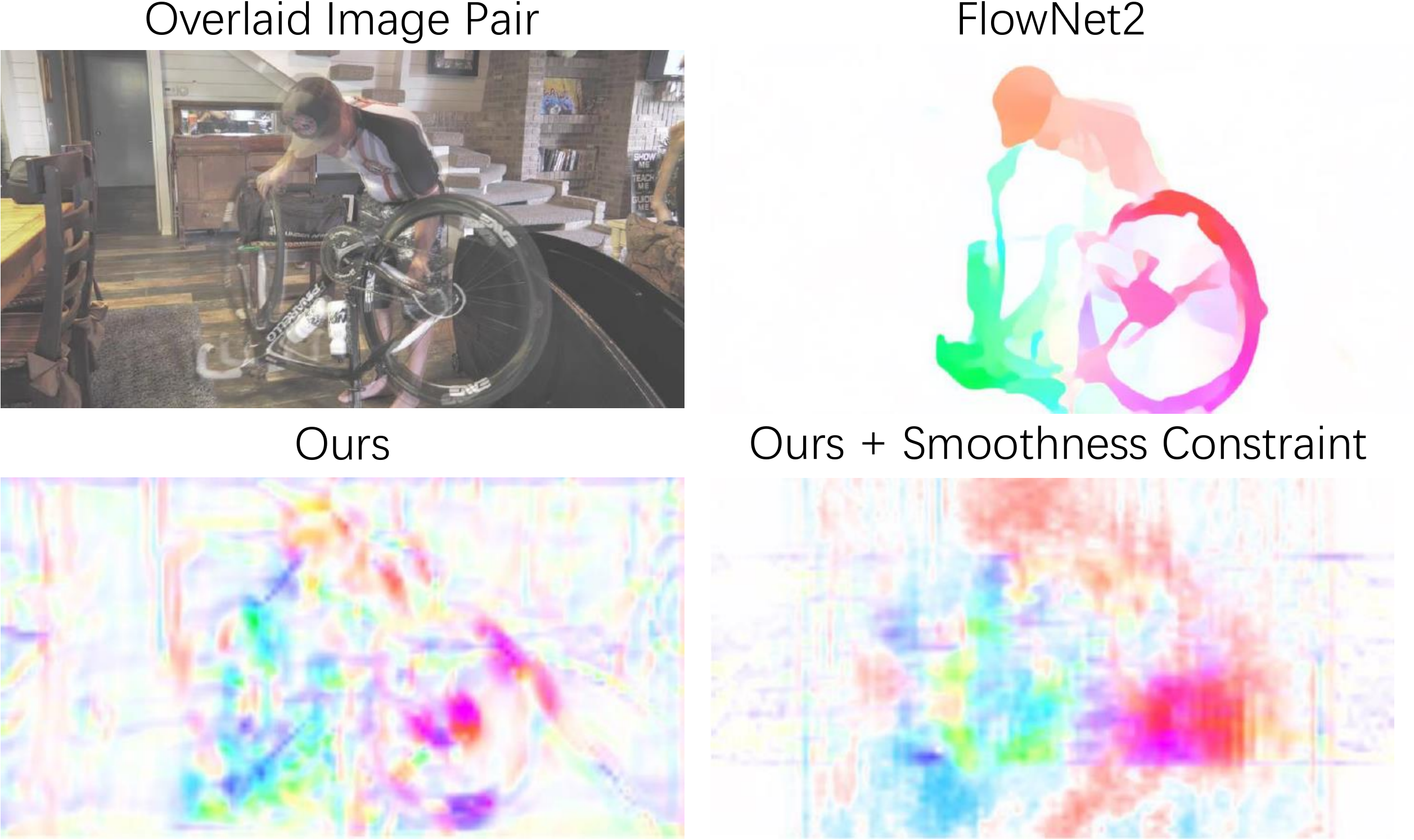}
	\caption{An example of optical flow computed from our model compared to FlowNet2.}
	\label{fig:flow}
	\vspace{-4pt}
\end{figure}

\vspace{2pt}
\textbf{Youtube-VOS.} We summarize the results on youtube-VOS validation set in Table~\ref{table:youtube}. Ours surpasses all prior works except STM~\cite{oh2019video}, which relies on heavy pretraining on a variety of segmentation datasets such as saliency detection and instance segmentation. Without pretraining, STM obtains a comparable result of $68.1\%$. We also test the generalization ability of our model trained on DAVIS train-val and test on Youtube-VOS val. The transferred model shows great generalization ability with an overall score of $67.4\%$

\vspace{2pt}
\textbf{Speed analysis.}
During tracking, we cache the appearance embeddings for a history up to 40 frames. 
Inference per frame thus only involves a feed-forward pass of the target frame through the base network, and an additional dot product of target embeddings to prior embeddings.
Computation is also constant of any number of objects.
This makes our algorithm extremely fast, with a runtime of $37$ frames per second on a single Titan Xp GPU.
Figure~\ref{fig:speed_g} compares current algorithms on their trade-off between speed and performance.
Ours is an order of magnitude faster than prior methods, while achieving 
the results comparable to state-of-the-art methods.

\subsection{Discussions}

\textbf{Temporal stability.}
\label{stable}
Temporal stability is often a desirable property in video object segmentation, as sharp inconsistencies may be disruptive to downstream video analysis.
However, temporal stability is typically not included as an evaluation criterion.
Here, we give qualitative examples showing the difference in temporal stability between
our model and the state-of-the-art PreMVOS~\cite{luiten2018premvos}.

In Figure~\ref{fig:stable}, we show examples of per-frame evaluation along video sequences. Although the state-of-the-art integrates various temporal smoothing modules, such as optical flow, merging and tracking, we observe the detection-based method to be prone to noise.
For example, objects are lost suddenly, or being tagged with a different identity.
Our method, on the other hand, makes temporally consistent predictions.

\vspace{2pt}
\textbf{Does our model learn optical flow?}
Our method learns a soft mechanism for associating pixels in the target frame with pixels in the history frames. 
This is similar to optical flow where hard correspondences are computed between pixels. 
We examine how much our learned model aligns with optical flow.

We take two adjacent two frames and calculate the optical flow from our model as $\Delta d_i = \sum_{j} s_{ij} \Delta d_{ij}$, where $s_{ij}$ is the normalized similarity,  and $\Delta d_{ij}$ is the displacement between $i, j$. Figure~\ref{fig:flow} shows an example visualization of the flow. Compared to the optical flow computed by FlowNet2~\cite{ilg2017flownet}, our flow makes sense on objects that would be segmented, but is much more noisy on the background. 
We have further added a spatial smoothness constraint on the computed optical flow for jointly learning the embeddings, as widely used for optical flow estimation~\cite{fischer2015flownet,janai2018unsupervised}.
We observe that the constraint smooths the optical flow on the background, but fails to regularize the model for tracking. 
Adding the term consistently hurts the performance of video object segmentation.

\section{Conclusion}

We present a simple approach to semi-supervised video object segmentation.
Our main insight is that much more unlabeled structure in the spatio-temporal volume can be exploited for video object segmentation.
Our model finds such structure via transductive inference.
The approach is learned end-to-end, without the need of additional modules, additional datasets, or dedicated architectural designs.
Our vanilla ResNet50 model achieves competitive performance  with a compelling speed of 37 frames per second.
We hope our model can serve as a solid baseline for future research.


{\small
\bibliographystyle{ieee}
\bibliography{egbib}
}

\end{document}